\documentclass{article}

\usepackage[final]{corl_2024} 

\title{EmbodiedRAG: Dynamic 3D Scene Graph Retrieval for Efficient and Scalable Robot Task Planning}

\author{
  Meghan Booker$^{\dagger, 1}$
  \And
  Grayson Byrd$^{1,2}$
  \And
  Bethany Kemp$^1$ 
  \AND
  \And Aurora Schmidt$^1$ \\
  \And Corban Rivera$^{1, 2}$ \\
}

\usepackage{xcolor}

\usepackage{graphicx}
\usepackage{subfigure}

\usepackage{wrapfig}
\usepackage{subcaption}

\usepackage{xspace}
\newcommand{\eRAG}{\textsc{EmbodiedRAG}\xspace}
\newcommand\blfootnote[1]{%
  \begingroup
  \renewcommand\thefootnote{}\footnote{#1}%
  \addtocounter{footnote}{-1}%
  \endgroup
}

\usepackage{titlesec}
\titlespacing*{\section}{0pt}{5pt}{5pt}
\titlespacing*{\subsection}{0pt}{3pt}{3pt}
\begin{document}
\makectitle

\blfootnote{$^\dagger$Corresponding Author: meghan.booker@jhuapl.edu. $^1$Johns Hopkins University Applied Physics Laboratory. $^2$Johns Hopkins University. 
}
\captionsetup[subfigure]{labelformat=empty}

\vspace{-24pt}
\begin{abstract}
    Recent advances in Large Language Models (LLMs) have helped facilitate exciting progress for robotic planning in real, open-world environments. 3D scene graphs (3DSGs) offer a promising environment representation for grounding such LLM-based planners as they are compact and semantically rich. However, as the robot’s environment scales (e.g., number of entities tracked) and the complexity of scene graph information increases (e.g., maintaining more attributes), providing a 3DSG as-is to an LLM-based planner quickly becomes infeasible due to input token count limits and attentional biases present in LLMs. 
    Inspired by the successes of Retrieval-Augmented Generation (RAG) methods that retrieve query-relevant document chunks for LLM question and answering, we adapt the paradigm for our embodied domain. Specifically, we propose a 3D scene subgraph retrieval framework, called \eRAG, that we augment an LLM-based planner with for executing natural language robotic tasks. Notably, our retrieved subgraphs adapt to changes in the environment as well as changes in task-relevancy as the robot executes its plan. We demonstrate \eRAG’s ability to significantly reduce input token counts (by an order of magnitude) and planning time (up to 70\% reduction in average time per planning step) while improving success rates on AI2Thor simulated household tasks with a single-arm, mobile manipulator. Additionally, we implement \eRAG on a quadruped with a manipulator to highlight the performance benefits for robot deployment at the edge in real environments.
\end{abstract}



\section{Introduction}
\label{sec:introduction}	
Exciting recent advances in 3D scene graphs (3DSGs) (e.g., \cite{armeni20193d, clio, gu2024conceptgraphs, rosinol20203d}) and large language model (LLM)-based planning (e.g., \cite{brohan2023can, rana2023sayplan, silver2024generalized, liu2023llm+, yao2022react,huang2022inner}) are facilitating the expansion of robot task planning and execution to rich, large-scale environments and to complex, long-time horizon tasks. In particular, 3D scene graphs allow for a robot to maintain a rich, semantic, yet compact representation of its operating environment which can then be serialized for use with an LLM-based planner. Thus, with language serving as a bridge, we can ground the LLM-based planner within the robot's operating environment for effective planning leveraging an LLM's strengths (e.g., common sense reasoning).

However, as the robot's environment scales (e.g., from rooms to houses to office buildings) or the complexity of entity information  increases (e.g., maintaining more attributes or detailed descriptions per entity), providing the 3DSG as-is to the LLM-based planner presents a couple of challenges. Almost immediately, the 3DSG will take up the token input limit for an LLM (e.g., GPT-4o has a 128k token limit). Even if input token count is not a limitation, reducing the input token count has the important potential to speed up planning time for the robot. Secondly, an LLM's attention capabilities are lacking; LLMs have been shown to be distracted by task-irrelevant information present in input prompts \cite{shi2023large, oh2023detrimental} as well as have positional bias to information provided at the beginning or ending of input prompts \cite{xiao2023efficient, liu2024lost}. Such limitations make it challenging for the LLM to identify important entities and environmental changes necessary for completing its task. Removing task-irrelevant information also removes associated uncertainty present in the 3DSG from the LLM potentially leading to more robust plans.

To mitigate these challenges, we take inspiration from the successes of Retrieval-Augmented Generation (RAG) methods (see \cite{gao2024retrievalaugmentedgenerationlargelanguage} for an overview) that retrieve task-relevant document chunks for LLM question and answering and adapt the paradigm to our embodied domain. Specifically, we seek to retrieve subgraphs from the robot's 3DSG that are \textit{task-relevant}, i.e., only contain entities, attributes, and relationships sufficient for completing a task. Importantly, the retrieved subgraph should adapt to changes in the environment and changes in task-relevancy. For example, if new, task-relevant information is added to the 3DSG, the subgraph provided to the LLM should incorporate this new information. Additionally, during replanning, the LLM-based planner may realize different entities could help complete the task due to what is present (or missing) in the robot's operating environment (i.e., the retrieved subgraph should incorporate feedback from planner).


\begin{figure}[t]
    \centering
    \includegraphics[width=0.99\linewidth]{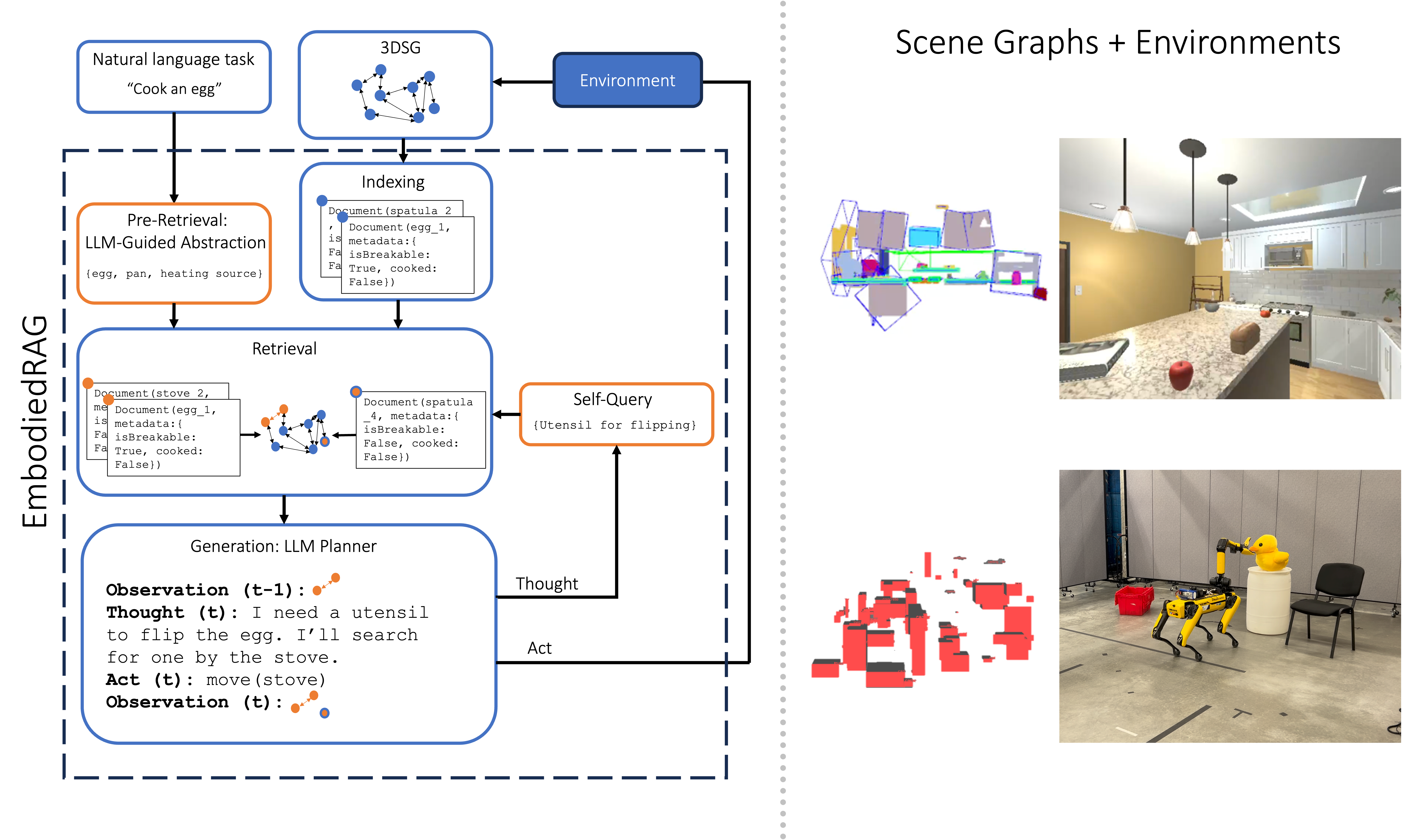}
    \caption{\footnotesize{\textbf{Left:} An illustration of the \eRAG framework. As a natural language task comes in, an LLM provides a quick abstraction for \textit{pre-retrieving} task-relevant entities and corresponding attributes without knowledge of the environment. As the robot begins executing the task, it builds a 3DSG representing the environment, nodes in the 3DSG are \textit{indexed} and updated in a vector store, and task-relevant entities are \textit{retrieved} and grounded to a subgraph. A ReAct-style LLM agent then \textit{generates} plans for the robot to execute using the subgraphs resulting in actions that affect the environment and thoughts on downstream actions that are used by an LLM-based \textit{self-query} mechanism to identify entities and attributes that are task-relevant to the particular operating environment and plan execution. 
    \textbf{Right:} Examples of scene graphs generated for the robot's operating environment in simulation (top) and hardware (bottom).}} 
    \label{fig:anchor}
\end{figure}

\textbf{Statement of Contributions.} Our primary contribution is a 3D scene subgraph retrieval framework, called \eRAG, that we augment an LLM-based planner with for robotic tasks. Our key innovation is adapting the RAG paradigm for the embodied domains with proposed methods for pre-retrieving task-relevant entities and attributes, indexing a 3DSG, retrieving and grounding to a 3DSG, generating plans, and incorporating planner feedback for retrieval. To our knowledge, this is the first work to extract relevant subgraphs for LLM-based planning that accommodates time-varying scene graphs (e.g., the 3DSG can be constructed on-the-fly), does not require an LLM to parse or manipulate the entire 3DSG for planning, and is agnostic to explicit 3DSG formatting (e.g., hierarchical, open-vocabulary, attribute types). We demonstrate \eRAG's ability to significantly reduce input token counts and planning time (up to 90\% cumulative token use and 70\% in average planning step time) while improving planning quality and success rates on AI2Thor simulated household tasks. We additionally implement \eRAG onboard a quadrupedal platform with a manipulator and compare against an agent using the full 3DSG (which often gets overwhelmed immediately by the token volume).



\section{Related Work}
\textbf{Task-Driven Representations} 
Task-driven representations aim to summarize the robot's state and environment \textit{sufficiently} for a given task (i.e., the robot has enough information to successfully complete the task). Humans are well studied for employing such representations (e.g., the gaze heuristic \cite{mcleod2003fielders, gigerenzer2007gut}) to acquire strong robustness to task-irrelevant distractors in the environment, improved planning efficiency for real-time decision making, and generalization to new tasks \cite{mccarthy2021connecting, ho2023rational}. Various approaches for robotics have sought to construct or learn task-driven representations using information bottlenecks \cite{Pacelli-RSS-20, clio, Soatto14}, minimizing the size complexity of the representation conditioned on the task \cite{o2017concise, booker2021learning}, or leveraging state abstraction theory for Markov Decision Processes \cite{li2006towards, zhang2021learning, booker2023switching}. The challenge with many of these approaches is scalability to complex environments and tasks; often knowledge about the environment or tasks is needed \textit{a priori}. In contrast, our framework  quickly generalizes to different natural language tasks and adapts to changes in task-relevancy online without needing information about the robot's operating environment \textit{a priori}.




\textbf{Retrieval-Augmented Generation.}
Retrieval-Augmented Generation (RAG) \cite{lewis2020retrieval} was first introduced to efficiently retrieve query-relevant document chunks to augment LLMs with domain-specific or up-to-date information in order to answer the query correctly. Since then many techniques and adaptations have been introduced such as pre-retrieval query rewriting \cite{zheng2023take}, iterative knowledge retrieval \cite{shao2023enhancing}, and extensions to knowledge graph structures \cite{edge2024local}. In contrast to the canonical RAG formulation, we are operating in an \textit{embodied} domain where a robot has the ability to change the information sources being retrieved (i.e., affect the environment), the information sources may be time-varying, and the information that is important to the original query may change as the robot executes a plan. 






\textbf{Planning with 3D Scene Graphs.}
Recent planning approaches using 3DSGs rely on text and image embeddings (e.g., SentenceBERT \cite{reimers2019sentence}, CLIP \cite{radford2021learning}) for retrieving relevant entities to a natural language task or sub-task \cite{gu2024conceptgraphs, clio, pmlr-v229-chang23b}. Gu et. al. \cite{gu2024conceptgraphs} show that using just CLIP embeddings misses out on capturing affordances or negations in the task and instead suggest using an LLM to extract an entity directly from the 3DSG; such an approach will become infeasible as the size of the 3DSG grows. Chang et. al. \cite{pmlr-v229-chang23b} propose entity retrieval through subgraph matching with embeddings. In general, these approaches place focus on the object localization problem (i.e., retrieve a singular entity related to the natural language task) and misses out on capturing multiple entities, specific state or attribute information, and relationships that could be necessary for more complex task planning (i.e., an LLM-based planner will be missing key environmental context). 

Some directions aim to find all the task-relevant entities with learning-based methods \cite{ploi2021} or graph pruning strategies \cite{taskography}. Most similar to our work is SayPlan \cite{rana2023sayplan} that seeks to provide an LLM-based planner a compressed 3DSG by allowing an LLM to expand and contract nodes in the 3DSG during planning. However, compression benefits are tied to the hierarchical structure of the provided 3DSG, and thus, misses out on significant compression at the lower levels (e.g., all objects within a room are probably not important for the task). Additionally, their approach assumes access to a pre-constructed 3DSG. Our work develops an approach for retrieving multiple task-relevant entities, entity-specific state or attribute information, and their relationships without requiring an LLM to parse a large 3DSG or access to the 3DSG \textit{a priori}.

\section{Problem Formulation}
Our goal is to design a framework for extracting subgraphs to augment an LLM-based robot planner with that (i) extracts minimal and task-relevant subgraphs in order to scale to large environments and accommodate LLM input token limits, (ii) adapts to time-varying 3DSGs (i.e., does not need a known 3DSG \textit{a priori}), and (iii) is agnostic to explicit 3DSG formatting (e.g., does not require a hierarchical structure or specific attributes). To formalize this, we pose the environment the robot is operating in $\mathcal{E}$ as a partially observable Markov Decision Process (POMDP) where at a state $s_t$ at time $t$, and provided a natural language task $l$, the robot executes action $a_t \in \mathcal{A}$ according to a policy $\pi$, and transitions to new state $s_{t+1}$. We assume the robot is constructing and maintaining a 3DSG semantically representing the observed state of $\mathcal{E}$,  $\mathcal{G}_t = \langle O_t, E_t \rangle$, where $O_t = \{o_j\}_{j=1\dots J}$ and $E_t = \{e_m\}_{m=1\dots M}$ are the sets of entities (e.g., objects) and edges respectively at a given time step. For each entity $o_j$, information about that entity may be stored as attributes $C^{attr}_{j,t} = \{c_{j, i}\}_{i=1 \dots I} $ (e.g., properties, action affordances, image embeddings, etc.), and similarly for edges, $C^{edge}_{m,t} = \{c_{m, f}\}_{f=1 \dots F} $ may capture inter-entity relationships (e.g., a caption, ``an egg is often in a fridge," or an attribute for a specific type of relationship, $o_1$ is on $o_2$). 

We additionally assume that policy $\pi$ is generated by an LLM provided with $\mathcal{G}_t$ and context window $x_t$, which contains the natural language task $l$ and any additional prompting $p$. An augmented context window $\tilde x_t := \{l, p, \mathcal{G}_{1}, a_1, \dots, \mathcal{G}_{t-1}, a_{t-1}\}$ contains past histories of LLM plans and may be given to the LLM. To provide the LLM-based planner with a subgraph in lieu of the full 3DSG, we need an extraction strategy $f: (\mathcal{G}_t, \tilde x_t) \mapsto \mathcal{G}'_t$ where $\mathcal{G}'_t = \langle  O'_t, E'_t \rangle$ is the extracted subgraph, i.e., $ O'_t \subseteq O_t$ and $ E'_t \subseteq E_t$. The challenge, then, lies in developing a good, \textit{time-varying} extraction strategy that provides a subgraph containing enough entities, entity-specific attributes, and relationships to successfully reason about the task at hand, $l$. We discuss our proposed approach, \eRAG, in the subsequent section.

\section{EmbodiedRAG}
\label{sec:eRAG}
To extract task-relevant subgraphs, our key insight is to connect our problem to the RAG paradigm. RAG frameworks typically have a document indexing phase for chunking information to be embedded into vector representations, a retrieval phase that brings up top relevant documents related to the user query, and then a generation phase where an LLM answers the query using the retrieved documents \cite{lewis2020retrieval, gao2024retrievalaugmentedgenerationlargelanguage}. More advanced RAG frameworks additionally use pre-retrieval phases to enhance the quality of the documents retrieved by using strategies such as query transformation or rewriting \cite{zheng2023take}. In the following subsections, we discuss how these RAG phases inspire our subgraph extraction strategy $f(\mathcal{G}_t, \tilde x_t)$ and how we incorporate feedback mechanisms for operating in real environments (see Figure \ref{fig:anchor}).



\subsection{Document Indexing a 3DSG}
In order to work in the RAG paradigm, we first need to connect the notion of documents to a 3DSG. Typically, an important consideration in RAG frameworks is deciding how best to chunk document content (e.g., sentences, paragraphs, pages, images). 3DSGs are already chunked into entities by construction, so a natural formulation is to treat each entity $o_j$ as a document to be retrieved. Specifically, we assign the object class or open-vocabulary label as a document's page content. Attributes $C^{attr}_{j,t}$ are assigned as the document metadata. Alternatively, attributes can be treated as sub-documents such that if they are retrieved, the parent document $o_j$ is retrieved. The latter would be preferable in cases where the attributes are not concise (e.g., long captions). Each document is then embedded in a vector store for efficient retrieval in downstream tasks. Open source, local vector stores such as Chroma \cite{chroma} and FAISS \cite{douze2024faiss} can be used in LLM integration frameworks such as LangChain \cite{langchain}.

\subsection{LLM-Guided Abstractions for Pre-Retrieval}
The next RAG component we take inspiration from is the pre-retrieval phase. For our setting, we want to avoid having an LLM parse or traverse an entire 3DSG, and at the start of robot deployment, the 3DSG might not be available yet. Thus, we want a strategy that enhances the quality of retrieved entities as knowledge about them becomes available. We propose generating an entity-oriented abstraction from the natural language task $l$ and the attributes that will be maintained in the 3DSG. Concretely, we ask an LLM prior to planning and any knowledge about the environment, i.e., no access to $\mathcal{G}_t$, to suggest a set of relevant entities $\{\hat o_r\}_{r=1\dots R}$ that would be useful for the natural language task $l$. If attributes are expected to be tracked in the 3DSG (a design choice known prior to robot deployment), then for each entity $\hat o_r$, we ask an LLM to suggest which attributes are important. Our pre-retrieval LLM prompts are provided in Appendix \ref{appendix:prompts}.

To illustrate the benefit of this pre-retrieval method, consider an in-home robot tasked with cooking an egg for a human. Intuitively the robot will at least need an egg, a cooking pan or container, and a heating source in order to complete this task. This quick abstraction can be inferred directly from the task description, can handle negations, affordances, and multiple entities, and does not require explicit knowledge about the home environment it is operating in (e.g., locations of objects, geometry of the kitchen, option of objects available to the robot, etc.). The entities in the abstraction may not even be present in the robot's operating environment but serve as a strong, initial direction for what should be retrieved, if they exist. If the 3DSG is expected to maintain attributes \{\texttt{temperature, breakable, toggleable}\}, then the only attributes relevant to an egg given the task are \{\texttt{temperature, breakable}\} while the attribute relevant to a pan is \{\texttt{temperature}\}.



\subsection{Subgraph Grounding for Retrieval
}
As the robot begins deployment in its operating environment $\mathcal{E}$, 3DSG construction begins and each entity is embedded as a document. To construct subgraph $\mathcal{G}'_t$, we first retrieve a thresholded, top-$k$ similarity documents to the pre-retrieved entities $\{\hat o_r\}$. The retrieved entities serve as entry points into our 3DSG, similar to how the local search component in graphRAG \cite{edge2024local} gains access to a knowledge graph. Next, we take the induced subgraph from these nodes to capture the relationships between the retrieved entities. If these are the only entities needed to successfully complete the task, then the LLM-based planner only needs to reason about their relationships. We introduce a feedback mechanism, described in Section \ref{sec: feedback}, to handle the scenarios when $O'_t$ does not contain all the entities needed to complete the task in environment $\mathcal{E}$ (e.g., occlusions, changes in task-relevancy, or missing entities or attributes due to a bad pre-retrieval). If there are attributes tied to pre-retrieved entity $\hat o_r$, then those attributes are applied to the set of retrieved entities for $\hat o_r$. We take this induced subgraph $\mathcal{G}'_t$ with filtered task-relevant attributes and serialize it to JSON format for the LLM-based planner.

\subsection{Plan Generation} 
The last main component of a RAG framework is the generation phase where the original query and retrieved documents are provided to an LLM in order to answer the query. Here, we want an LLM to generate actions $a_t$ instead of answers to a query, i.e., serve as our policy $\pi(a_t|\mathcal{G}'_t, \tilde x_t)$. To do so, we leverage a ReAct-style LLM agent \cite{yao2022react} that is designed to generate actions given the history of past actions and observations. Notably, the action space $\mathcal{A}$ is augmented with language to provide \textit{thoughts} $\tau_t$ about reasoning and future actions without affecting the environment. Thus, at each step $t$, we provide the ReAct-style LLM agent with its context history $\tilde x_t = \{l, p, \mathcal{G}'_{1}, (a_1, \tau_1), \dots, \mathcal{G}'_{t-1}, (a_{t-1}, \tau_{t-1})\}$ and an updated subgraph $\mathcal{G}'_t$ as the observation. 

\subsection{Incorporating Feedback}
\label{sec: feedback}

Up until now, the retrieval process has been open-loop and following a canonical RAG framework. However, the augmented actions, $(a_t, \tau_t)$, of the LLM-based planner have the ability to close feedback loops and affect components in the \eRAG framework in two ways. Intuitively, action $a_t$ directly affects the environment, which in turn, updates the documents or adds new documents. This feedback loop is critical for the LLM-based planner to verify that its action completed as intended (e.g., seeing attribute \texttt{temperature} increase for the egg if the LLM cracked the egg on the pan) or show newly discovered task-relevant entities to the planner (e.g., the LLM-based planner was searching for an egg in the first place and the egg was recently populated in the full 3DSG).

The second feedback component is via the thoughts $\tau_t$. Although thoughts do not affect the environment, our key insight is to use them to affect our retrieval mechanism as they provide insight into the type of information (e.g., specific types of objects or attributes) the LLM-based planner is currently making, would like to make, or plans to make decisions about. To process the thoughts $\tau_t$, we use 
\begin{wrapfigure}{r}{0.4\linewidth} 
    \centering
    \vspace{-8mm}
    \begin{subfigure}{}
        \centering
        \includegraphics[width=0.99\linewidth]{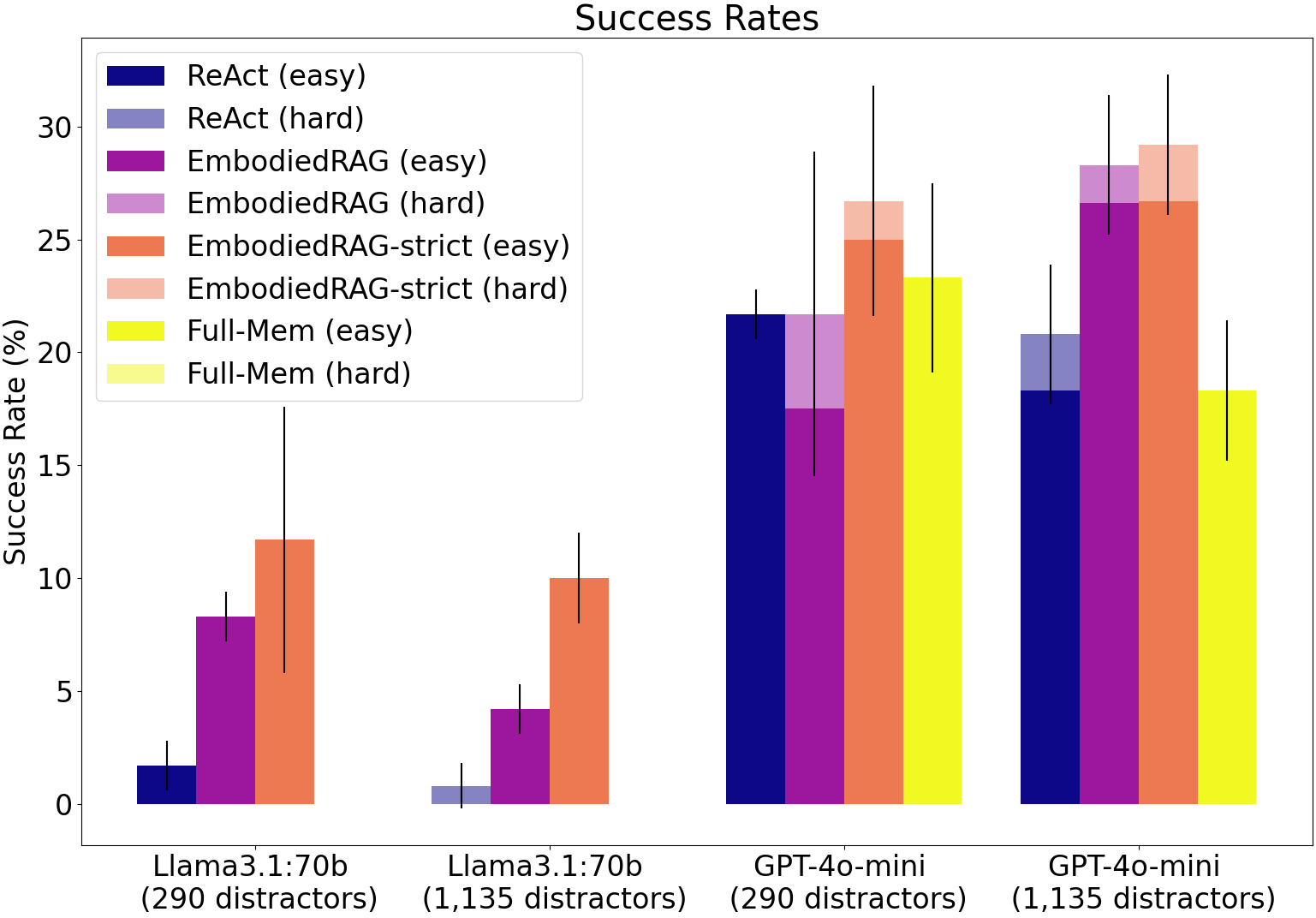}
    \end{subfigure}
    \vspace{-4mm}
    \begin{subfigure}{}
        \centering
        \includegraphics[width=0.99\linewidth]{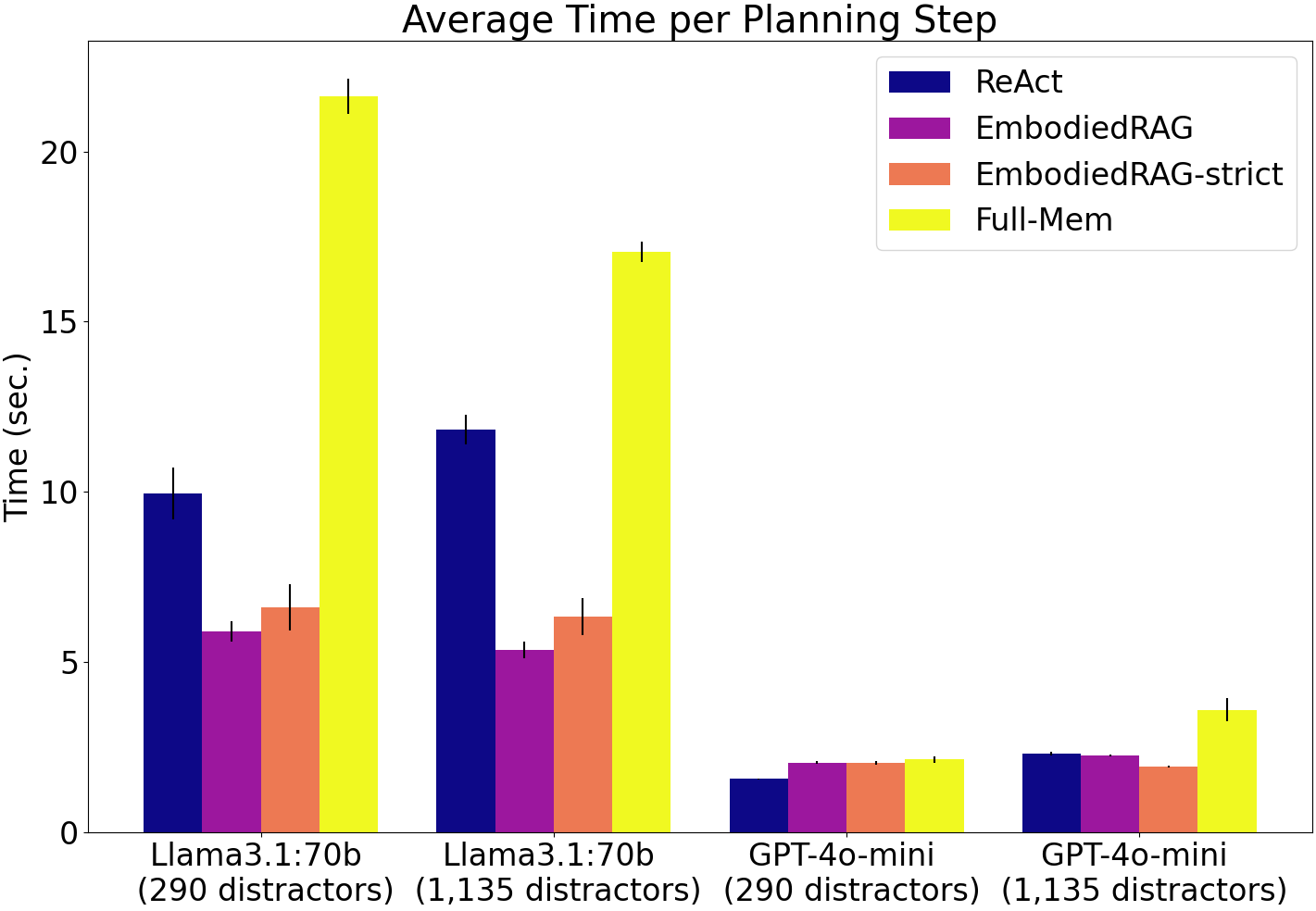}
    \end{subfigure}
    \vspace{-4mm}
    \begin{subfigure}{}
        \centering
        \includegraphics[width=0.99\linewidth]{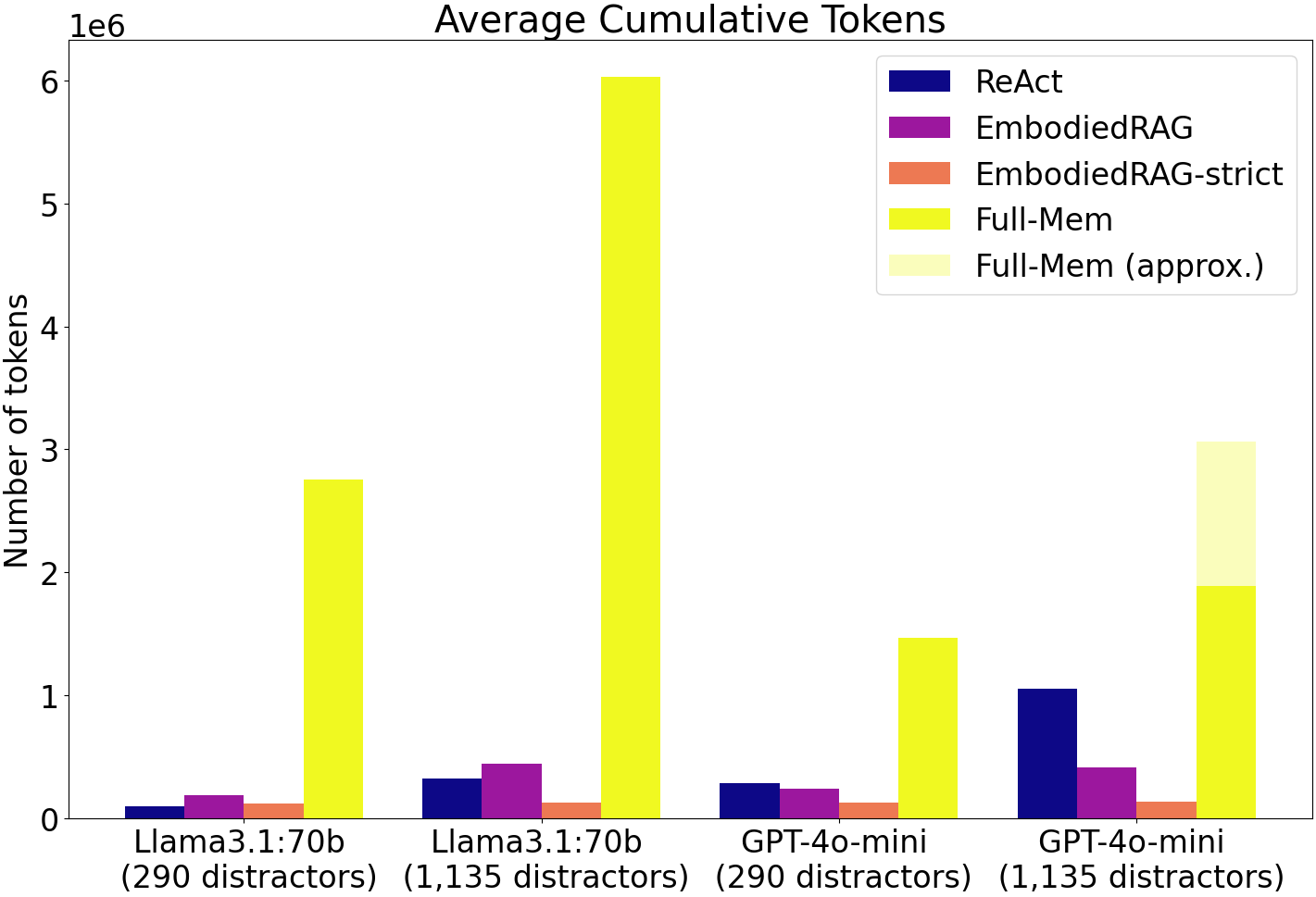}
    \end{subfigure}
    
    \caption{\footnotesize{Average performance of agents evaluated across three experiments with a varying number of distractor objects present in the 3DSG (either 290 or 1,135 distractor objects). Each experiment consists of 40 different tasks (20 easy, 20 hard) in five AI2Thor kitchen scenes. \textbf{(top)} Success rates. Darker and lighter colors reflect the proportion of successful easy and hard tasks respectively. Error bars (black lines) are the standard deviation of the total success rates across the three experiments. \textbf{(middle)} Average time in seconds per planning step. \textbf{(bottom)} Average cumulative tokens used to represent the 3DSG to the LLM-based planner. Note for \texttt{GPT-4o-mini} with 1,135 distractor objects, an average of 15 tasks per experiment did not complete due to input token count limits. Approximated token count is shown in light yellow using the average cumulative tokens used per task.}}
    \label{fig:distractors}
    \vspace{-16mm}
\end{wrapfigure}the self-query method for RAG (e.g., \cite{langchain-query}) that uses an LLM to generate a structured query from the original prompt in order to identify important semantics and perform metadata filtering. Instead of passing in the original task, we pass in the thought to identify key entities and their attributes (i.e., metadata to filter for). For example, when cooking the egg, the LLM-based planner may realize it needs a utensil to flip the egg, an entity not in the original pre-retrieval abstraction. The self-query mechanism will then propose retrieving an entity related to ``utensil for flipping." At retrieval time, entities such as a spatula or turner will be added to $\mathcal{G}'_t$ if they exist in $\mathcal{G}_t$. This type of feedback can also help retrieve entities related to occlusions or obstacles that become relevant as the task is executed in a specific environment as well as change the metadata as it becomes relevant. We show examples of additional retrieved entities and metadata in the subsequent section.

\section{Simulated Examples}
\label{sec:sim}

We evaluate the efficacy of \eRAG across simulated household tasks in four AI2Thor simulator \cite{kolve2017ai2} kitchen scenes (Figure \ref{fig:ai2thor}). Tasks range from 20 easy\footnote{Our easy tasks are simple search tasks that clearly specify which objects the agent needs to interact with.} (e.g., ``put the tea kettle on the stove") to 20 hard\footnote{Our hard tasks include multiple instructions or ambiguous descriptions that require common-sense knowledge to complete.} tasks (e.g., ``cook an egg" and ``place all kitchen utensils that are not knives away in drawers"). A single-arm, mobile manipulator has 40 iterations to complete the task and can perform object-oriented actions such as \texttt{pickup(object)} or \texttt{placeon(object)}. See Appendix \ref{appendix:sim} for the complete lists of tasks and actions.

In particular, we investigate the efficacy of \eRAG on input token count, planning efficiency, and success rate under two different scenarios: (i) presence of task-irrelevant, distractor entities and (ii) presence of many attributes.  
For each scenario, we compare our approach against two baselines. The first baseline is a simple \textbf{ReAct} agent \cite{yao2022react} that acquires its knowledge about the environment through an explicit action request, \texttt{getdiscoveredobjects()}\footnote{Calling this action is requesting access to the full 3DSG.}.  The second baseline, \textbf{Full-Mem}, is provided the full 3DSG $\mathcal{G}_t$ as an observation at each planning time step. Our approach, \textbf{\eRAG}, is only provided the retrieved subgraph $\mathcal{G}'_t$ as an observation but may get access to $\mathcal{G}_t$ with an explicit request. We additionally introduce \textbf{\eRAG-strict}, which does not have access to $\mathcal{G}_t$ at any time.

\subsection{Distractor Entities}
To emulate large environments as well as uncertainty around object existence that could occur during 3DSG construction, we investigate the impact of non-existent, distractor objects present in the 3DSG. In order to scale to a large number and avoid exceeding input token limits, we present only the set of entities in $\mathcal{G}_t$ (i.e., no attributes or edges) to the LLM-based planner. We evaluate the aforementioned agents with 290 and 1,135 distractor objects and show the results for two different LLMs (\texttt{llama3.1:70b}\footnote{\texttt{llama3.1:70b} was run locally using a 16-core CPU and two NVIDIA GeForce RTX 4090s.} and \texttt{GPT-4o-mini}) in Figure \ref{fig:distractors}.

Overall, \eRAG and \eRAG-strict demonstrate several important improvements compared to the baselines. Both are able to consistently complete several hard tasks with \texttt{GPT-4o-mini} and generally maintain success rate while significantly reducing average time per planning step and cumulative token use. Additionally, \eRAG-strict's performance across all three metrics does not degrade as the number of distractor objects increases while all of the other agents are impacted by the increase. This highlights the scalability of using our approach with an LLM-based planner.

\begin{table*}[t]
\begin{minipage} [t]{1.0\linewidth}
    \begin{center}
\begin{tabular}{ |l|c|p{2cm}|c|p{1.3cm}|}
 \hline
 \textbf{Agent} & \textbf{Success Rate} & \textbf{Avg. Time per Step (sec.)} &  \textbf{Avg. Cum. Tokens} & \textbf{Full 3DSG Requests}\\ 
 \hline
 \hline
 \multicolumn{5}{|c|}{\textbf{\textit{Simulation: }} \texttt{GPT-4o-mini}}
  \\
 \hline
 \hline
 ReAct & 0.125 $\pm$ 0.041 & 2.145 $\pm$ 0.055 & 1,112,598 & 125 \\ 
 Full-Mem & 0.066 $\pm$ 0.012 & 3.106 $\pm$ 0.359 & 1,297,389/2,515,975\footnote{The estimated cumulative token count using avg. token count per trial for trials that hit token input limits.} & 12\\ 

 \eRAG & 0.092 $\pm$ 0.024 & 2.065 $\pm$ 0.022 & 758,305 & 48\\
 \eRAG-strict & 0.133 $\pm$ 0.012 & \textbf{1.759} $\pm$ \textbf{0.025} & \textbf{423,442} & 0\\
 \eRAG-feedback & \textbf{0.150} $\pm$ \textbf{0.020} & 3.457 $\pm$ 0.098 & 501,611 & 0 \\
 \hline
\end{tabular}
\end{center}
\vspace{-2mm}
\caption{ \textup{Average performance of agents across three experiments (120 total trials) using 3DSG format with 28 attributes per entity. Each experiment consists of 40 different tasks in five AI2Thor kitchen scenes. }}


\label{tab:graph_results}
\end{minipage}
\end{table*}

\subsection{Full 3DSG with Distractor Attributes}

Next, we run the same tasks but construct the 3DSG, provided in JSON format to the LLM-based planner, with 28 possible attributes and edges between entities representing spatial and control relationships. A complete list of attributes populated for every entity is provided in Appendix \ref{appendix:attrs}. We additionally implement \textbf{\eRAG-feedback} to incorporate the self-query retrieval mechanism discussed in Section \ref{sec: feedback}. Results are shown in Table \ref{tab:graph_results} under Simulation for \texttt{GPT-4o-mini}\footnote{\texttt{llama3.1:70b} was unable to successfully complete the tasks for this richer 3DSG format.}. 

Notably, \eRAG-strict performs the best across the planning time and cumulative tokens without needing access to the complete 3DSG. The ReAct agent has a comparable success rate but uses over 2.5 times the amount of tokens and more time to reason about the observations. Incorporating the self-query retrieval improves the success rate of \eRAG-strict while still keeping token counts low\footnote{The token counts for the self-query retrieval are not included as these are not shown to the LLM-based planner. Affect of the self-query retrieval are included in planning time.} demonstrating the importance of feedback in the subgraph retrieval process. In particular, the agent was able to gain access to objects and attributes that were not originally task-relevant but became so during replanning in the agent's operating environment. Self-query examples are shown in Appendix \ref{appendix:self-query}. 
\vspace{-8pt}

\section{Hardware Example}
\label{sec:hardware}
Additionally, we demonstrate the efficacy of \eRAG running on a quadrupedal robot in a real environment. We equip Boston Dynamic's Spot with an NVIDIA Jetson AGX Orin and a Zed 2 RGB-D camera in order to perform 3DSG construction using ConceptGraphs \cite{gu2024conceptgraphs}. We also run the LLM-based planner using \texttt{llama3.1:8b} onboard. Similar to the easy tasks in Section \ref{sec:sim}, we task the Spot robot with mobile manipulation tasks (see Figure \ref{fig:anchor} bottom, right) under two different environment sizes. For the first case, the robot's 3DSG only contains a few entities that are visible immediately (small environment), and in the second case, we have the robot map a large lab and office space (large environment). We run 5 trials for each agent (Full-Mem and \eRAG-strict) in each environment size for 20 total trials. Results are shown in Figure \ref{fig:failure} and Table \ref{tab:hardware_planning}.

\begin{minipage}[h]{0.5\textwidth}
\vspace{6mm}
    \centering
    \includegraphics[width=\textwidth]{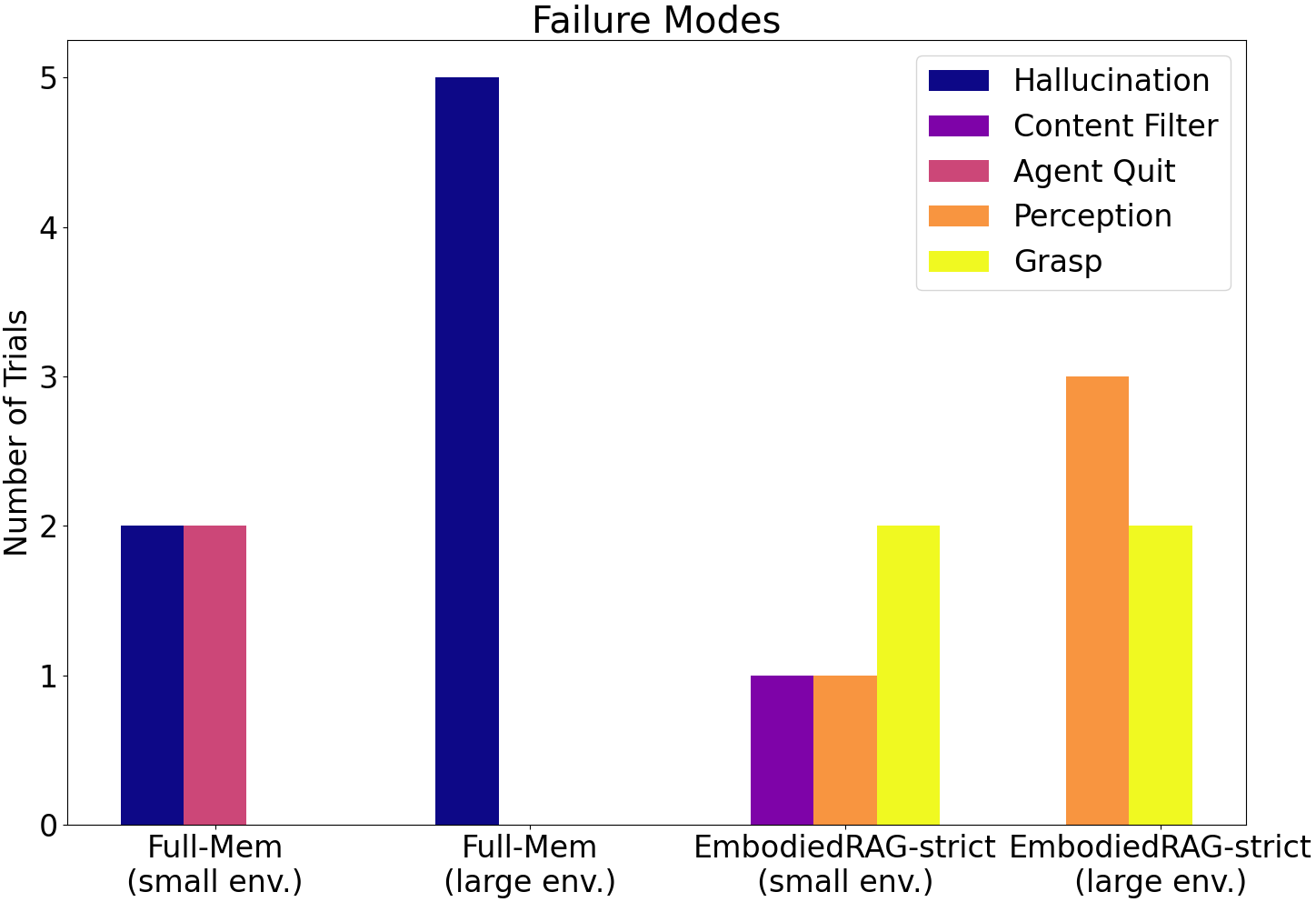}
    \captionof{figure}{Modes of failure. }
    \label{fig:failure}
\end{minipage}%
\hfill
\begin{minipage}[t]{0.45\textwidth}
\begin{tabular}{ |p{2.5cm}|p{1.1cm}|p{1.6cm}|}
 \hline
 \textbf{Agent} & \textbf{Steps to Failure} & \textbf{Avg. Time per Step (sec.)} \\ 
 \hline
 \hline
 \multicolumn{3}{|c|}{\textbf{\textit{Hardware: }} \texttt{llama3.1:8b}}
  \\
 \hline
 \hline
 Full-Mem (small) & 4.5 & 17.117  \\ 
 Full-Mem (large) & 1 &  \hspace{2mm} ---\footnote{No average time to report since Full-Mem did not get past the first planning step.} \\ 

 \eRAG-strict (small) & 4 & 13.101 \\
 \eRAG-strict (large) & 5.25 & 17.843 \\
 \hline
 \end{tabular}
 \captionof{table}{Planning metrics.}
 \label{tab:hardware_planning}
\end{minipage}

Both agents successfully completed one of 10 trials while in the small environment. However, there are significant differences in the quality of the plans provided, particularly regarding the failure modes (Figure \ref{fig:failure}; see Appendix \ref{appendix:hardwarelogs} for more details and examples). When switching from the smaller environment to the larger, Full-Mem more than doubled the amount of hallucinations and went from planning for multiple steps to only completing one before catastrophically hallucinating. Meanwhile \eRAG-strict did not hallucinate, and instead experienced an increase in perception errors as a result of mapping a larger space. Importantly, the increase in environment size did not impact the agent's ability to plan and attempt to recover as shown by a similar number of steps taken before failure in both environment sizes.


\section{Conclusion}
\label{sec:conclusion}

We developed and evaluated \eRAG to address the challenges of efficiently integrating 3D scene graphs with LLMs for robot task planning in dynamic, open-world environments. Notably, our approach does not require a 3DSG \textit{a priori} and it avoids providing an LLM-based planner the entire 3DSG, leading to faster planning generation and improved success rates as compared to a ReAct-only agent and one that receives the full 3DSG. Pivotal to our approach are the feedback components to aid in planning recovery during deployment. Actions suggested by the LLM-based planner affect the environment while the planner's natural language thoughts affect additional, task-relevant retrievals. 

\textbf{Limitations and Future Work.} The retrieved entities by \eRAG are dependent on those that are added to the 3DSG, and thus the performance can be affected by uncertainty involving perception (e.g., multiple entities representing the same object will result in multiple entities retrieved) and localization (e.g., the robot could move to the wrong location relative to a retrieved entity). Future directions can include expanding to multimodal retrieval techniques --- an avenue our framework is equipped to extend to. Our framework could also benefit from faster, structured queries for feedback; the self-query mechanism used in this work almost doubled the time per planning step despite improved success rates.  In particular, we hope our framework promotes further research in pre-retrieval, retrieval, and self-query components tailored for the embodied domain. 

\acknowledgments{The authors are thankful for the invaluable feedback and input provided by Emma Holmes, David Handelman, William Paul, and Tyler Feldman. This research was partially sponsored by Independent Research \& Development (IR\&D) program funds in the Research and Exploratory Development Department (REDD) at the Johns Hopkins University Applied Physics Laboratory. Additionally, this research was sponsored by the Army Research Laboratory and was accomplished under Cooperative Agreement Number W911NF-21-2-0211. The views and conclusions contained in this document are those of the authors and should not be interpreted as representing the official policies, either expressed or implied, of the Army Research Office or the U.S. Government. The U.S. Government is authorized to reproduce and distribute reprints for Government purposes notwithstanding any copyright notation herein. DISTRIBUTION A. Approved for public release; distribution unlimited.
}


\bibliography{bibliography}  

\appendix

\section{Prompts}
\label{appendix:prompts}
\subsection{Pre-Retrieval Prompt}
Here is the prompt we use to generate an entity-oriented abstraction for a task and without knowledge of what the ``environment" is.

\texttt{\{task\} What objects in the environment are important to consider for a robot agent completing this task? Strictly return a comma separated list of objects only.}

Once the LLM-guided abstraction is generated, we ask for each object proposed what attributes will be relevant to the task. Attributes are known prior to task planning and execution based on 3DSG implementation being used. 

\texttt{Which of the following attributes about a \{object\} are important to consider for task: \{task\}?
        Possible attributes:  \{attributes\}.
        Strictly return a comma separated list of the attributes only.}

\subsection{Self-Query Prompt}
We use the prompt provided by Langchain for self-query \cite{langchain-query} with a JSON formatted list of attributes.

\section{Simulation Details}
\label{appendix:sim}
\subsection{Tasks}

\textbf{Easy Tasks.}
\begin{enumerate}
    \item Pick up the pot that is on the counter top and place it on the shelf
    \item Pick up the credit card that is on the counter top and place it in the drawer
    \item Pick up the vase that is on the shelf and place it in the cabinet
    \item Pick up the lettuce that is on the counter top and place it in the garbage can
    \item Pick up the apple that is on the counter top and place it in the pot that is on the counter top
    \item Pick up the pepper shaker that is on the counter top and place it on the shelf
    \item Pick up the potato that is on the counter top and place it in the garbage can
    \item Pick up the cup that is in the sink and place it in the microwave that is on the counter top
    \item Pick up the spatula that is on the counter top and place it in the bowl that is on the counter top
    \item Put the pepper shaker found on the counter top in a drawer
    \item Pick up the fork that is on the counter top and place it in the cup that is on the counter top
    \item Pick up the potato that is on the counter top and place it in the fridge
    \item Put a tomato on a plate
    \item Pick up the cell phone that is on the shelf and place it on the counter top
    \item Put the credit card on the counter top in a drawer
    \item Take the bread from the dining table and place it on a counter
    \item Pick up the mug that is in the coffee machine and place it in the sink basin
    \item pick up the dish sponge that is on the counter top and place it in the sink basin
    \item Pick up the lettuce that is on the dining table and place it on the counter top
    \item Pick up the kettle that is in the cabinet and place it in the sink basin
\end{enumerate}

\textbf{Hard Tasks.}
\begin{enumerate}
    \item Cook an egg
    \item Water the plant
    \item Place a fork in a pot and a spoon in the sink
    \item Boil a potato and fry an egg. Leave the cooked food inside what they were cooked in
    \item Cook a potato without using the stove. Leave the potato inside what it was cooked in
    pick up the pepper shaker that is on the counter top and place it on the shelf
    \item Chill the tomato
    \item Water the plant and remove all objects from the fridge
    \item Put the bowl away in a cabinet and put the mug in the sink
    \item The apple in the fridge is rotten. Dispose of it
    \item Store the bread and potato. Do not store any of them on a shelf or in a fridge
    \item Fill the cup, mug, and pot with water
    \item Place the lettuce, tomato, and potato in the sink
    \item I need to run errands tomorrow. To do this, I will need something to pay with and something to use to navigate. Place these two items inside a drawer in preparation for my trip
    \item Remove all items from the sink
    \item Place all kitchen utensils that are not knives away in drawers
    \item Boil some water for tea
    \item Find all eating utensils besides knives and place them on the dining room table
    \item Clear off the dining room table
    \item Put the lettuce away where it will stay fresh
    \item Put a fruit in the fridge and close the fridge door. Next, place a tomato in a cabinet
\end{enumerate}

\begin{figure}[h]
      \centering
      \includegraphics[width=0.98\linewidth]{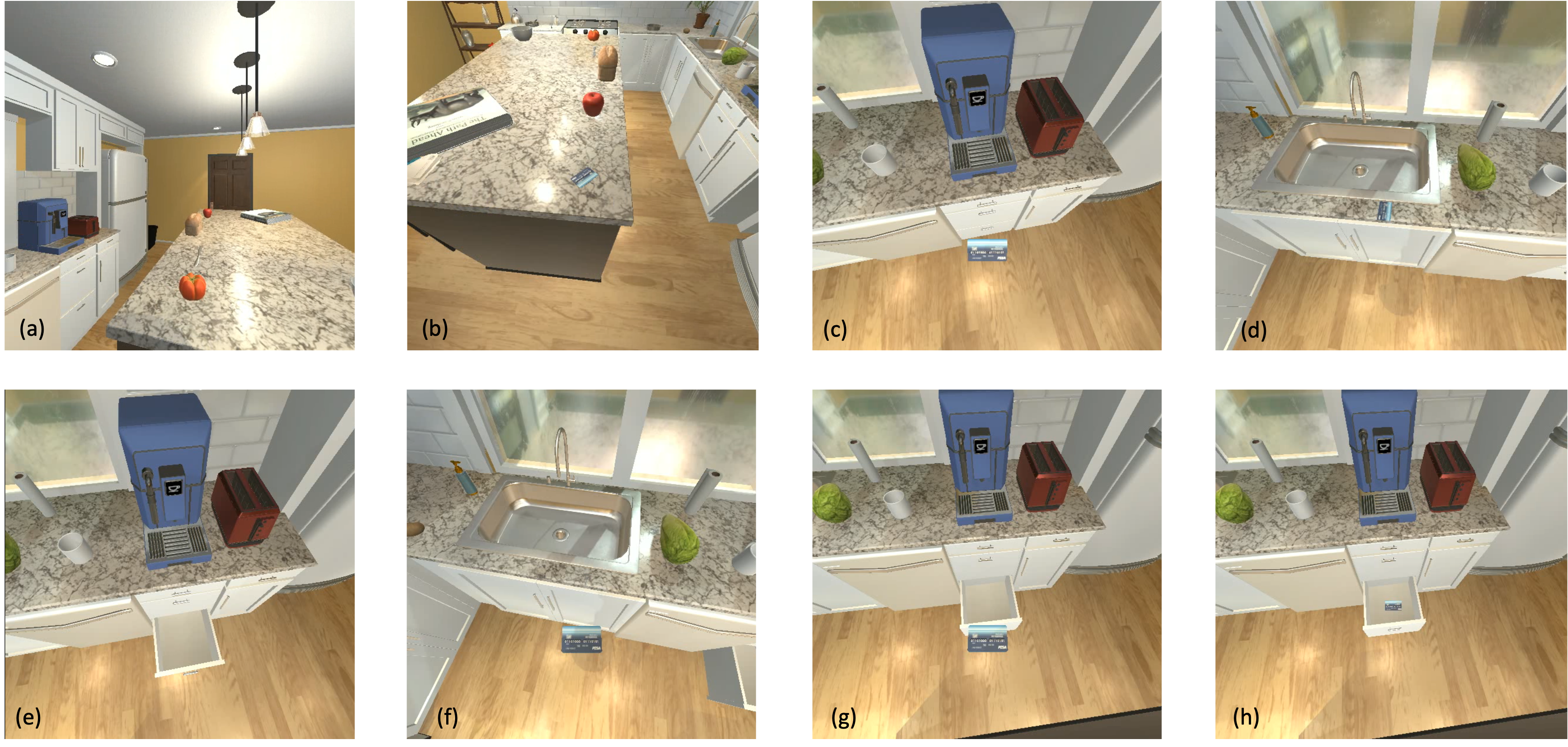}
      \caption{Example task completion in AI2Thor by \eRAG-feedback agent. \emph{Task: Pick up the credit card that is on the counter top and place it in the drawer.} Prior to task execution, the agent generates abstraction \texttt{\{credit card, counter top, drawer\}}. (a) the agent starts in the kitchen, (b) after some exploration, the credit card is visible to the agent, added to the 3DSG, and relayed to the agent via retrieval, (c) the agent takes the credit card to the kitchen drawer, but it's closed, (d) the agent adapts by placing the card by the kitchen sink, before (e) moving back to the drawer and opening it.  (f) the agent moves back to the sink to pick up the card again where the self-query mechanism includes additional attribute information for the credit card \texttt{isPickedUp}, before (g) moving back to the open drawer, and (h) placing the credit card in the drawer to complete the task. }
      \label{fig:ai2thor}
\end{figure}

We show an example task completion using \eRAG-feedback in Figure \ref{fig:ai2thor}.

\subsection{Action Tooling}
The actions provided to the LLM-based planner for the AI2Thor simulations in Section \ref{sec:sim}. Note \eRAG-strict and \eRAG-feedback were not provided the \texttt{getdiscoveredobjects()} as this returns the full 3DSG.

\begin{itemize}
    \item \texttt{randomlyexplore()}
    \item \texttt{getdiscoveredobjects()}
    \item \texttt{getvisibleobjects()}
    \item \texttt{moveto(object)}
    \item \texttt{inspect(object)}
    \item \texttt{pickup(object)}
    \item \texttt{placeon(object)}
    \item \texttt{open(object)}
    \item \texttt{close(object)}
    \item \texttt{toggleon(object)}
    \item \texttt{toggleoff(object)}
    \item \texttt{search(object)}
    \item \texttt{fillheldobjectwithwater()}
    \item \texttt{pourwaterinto(object)}
    \item \texttt{adjustpositioning()}
    
\end{itemize}

\subsection{3DSG Attributes}
\label{appendix:attrs}
The attributes added for an entity in the AI2Thor simulations in Section \ref{sec:sim}. 
\begin{itemize}
    \item \texttt{visible}
    \item \texttt{isInteractable}
    \item \texttt{toggleable}
    \item \texttt{isToggled}
    \item \texttt{breakable}
    \item \texttt{isBroken}
    \item \texttt{canFillWithLiquid}
    \item \texttt{isFilledWithLiquid}
    \item \texttt{fillLiquid}
    \item \texttt{dirtyable}
    \item \texttt{isDirty}
    \item \texttt{canBeUsedUp}
    \item \texttt{isUsedUp}
    \item \texttt{cookable}
    \item \texttt{isCooked}
    \item \texttt{temperature}
    \item \texttt{isHeatSource}
    \item \texttt{isColdSource}
    \item \texttt{sliceable}
    \item \texttt{isSliced}
    \item \texttt{openable}
    \item \texttt{isOpen}
    \item \texttt{openness}
    \item \texttt{pickupable}
    \item \texttt{isPickedUp}
    \item \texttt{moveable}
    \item \texttt{salientMaterials}
    \item \texttt{distance}
\end{itemize}

\subsection{Self-Query Examples}
\label{appendix:self-query}
\begin{figure}[h]
      \centering
      \includegraphics[width=0.98\linewidth]{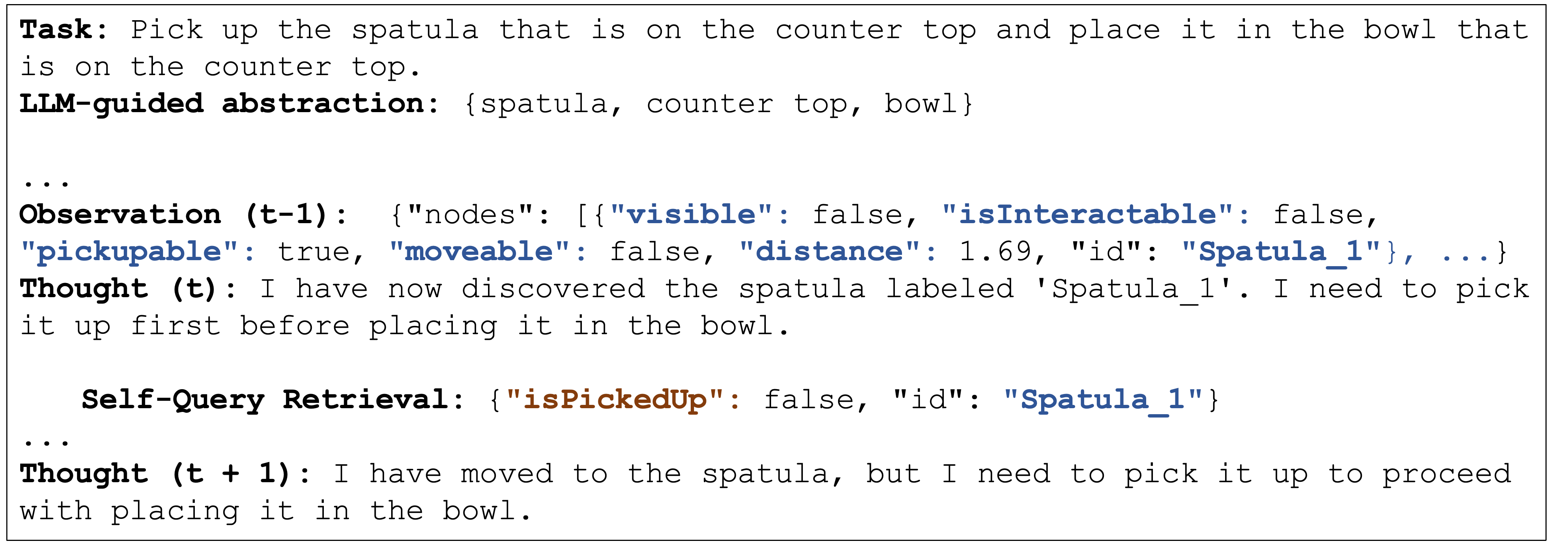}
      \caption{Example of the self-query retrieval adding task-relevant state information (orange) for the object \texttt{Spatula\_1} that was already in the retrieved subgraph (blue).}
      \label{fig:appendix_state}
\end{figure}

\begin{figure}[h]
      \centering
      \includegraphics[width=0.98\linewidth]{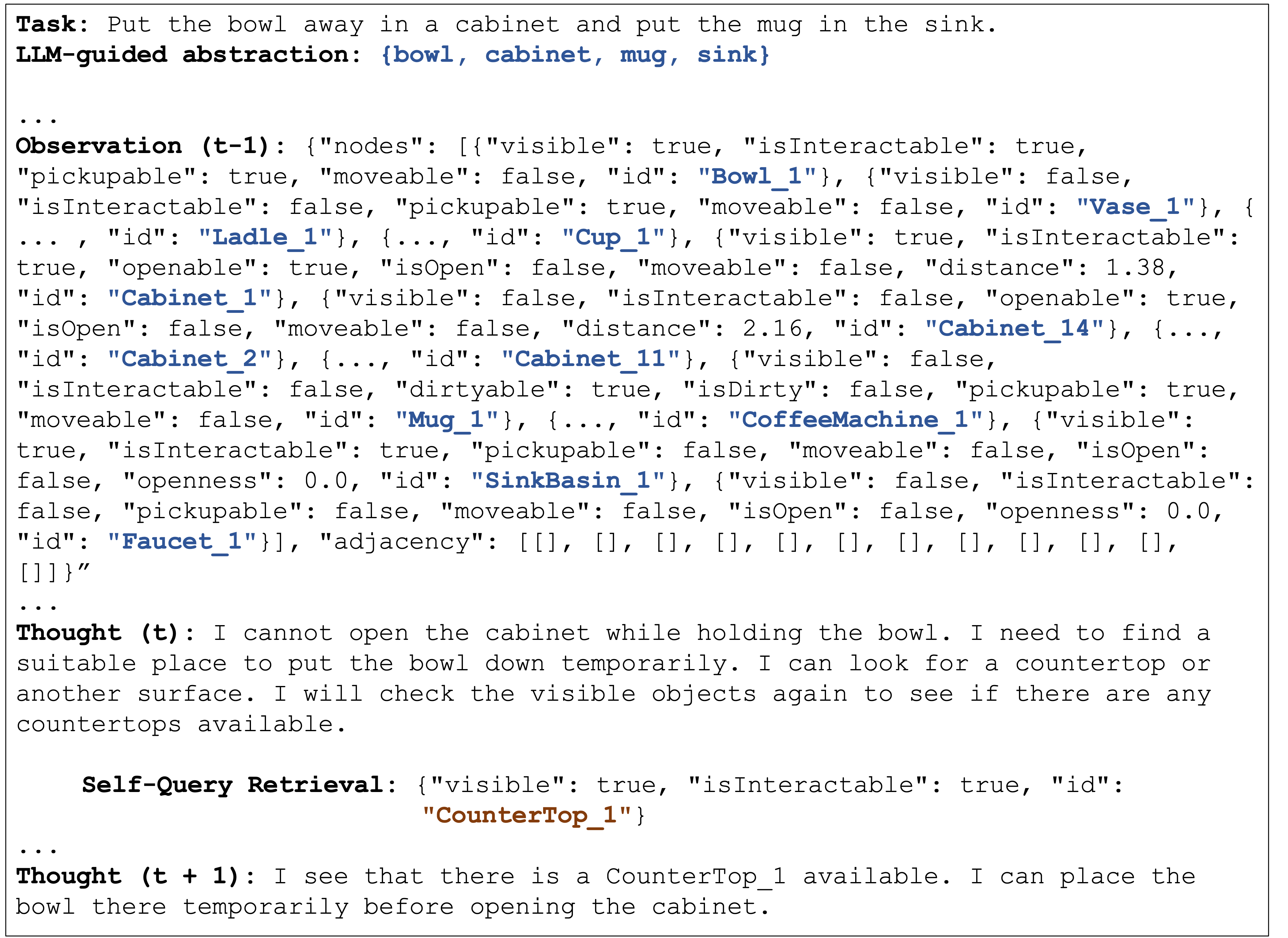}
      \caption{Example of the self-query retrieval adding task-relevant entities (orange) not originally proposed to be needed for the task (blue) but became relevant during replanning at Thought (t).}
      \label{fig:appendix_entities}
\end{figure}

We provide two examples of the self-query retrieval (used by \eRAG-feedback) adding task-relevant information that was not known at the start of task but became relevant including additional state information (Figure \ref{fig:appendix_state}) and additional entities (Figure \ref{fig:appendix_entities}).

\section{Hardware Details}
\label{appendix:hardware}

\subsection{Action Tooling}
The actions provided to the LLM-based planner for the hardware implementation in Section \ref{sec:hardware}. 

\begin{itemize}
    \item \texttt{locatebyID(object)}
    \item \texttt{movenear(pose)}
    \item \texttt{moverelative(pose)}
    \item \texttt{speak(string)}
    \item \texttt{grab(object)}
    \item \texttt{groundviewandgrab(object)}
    \item \texttt{release()}
    \item \texttt{robotposition()}
    \item \texttt{gettime()}
    \item \texttt{sit()}
    \item \texttt{stand()}
    
\end{itemize}

\subsection{Examples}
\label{appendix:hardwarelogs}

\subsubsection{Failure Modes}
Failure modes included hallucination by the agent, the agent refusing to manipulate the environment out of safety (content filter), the agent ending the task due to repeated action failures and not suggesting alternative actions (agent quit), poor entity locations that resulted in halting the trial to avoid crashes (perception), and trial halts from poor grasp executions (grasp).

\subsubsection{Example Hallucination}Example hallucination from Full-Mem agent in hardware after one planning step in the large environment.

\texttt{It looks like you've provided a list of items, followed by a seemingly unrelated thought. However, I think I see what's going on here!
The first part of the text appears to be a series of words related to objects, categorizing them with numbers (e.g., ladder$\_$56, truss$\_$178, etc.). This might be some sort of data or inventory system.
The Thought: section seems to be... just a thought! It's not directly related to the list of objects. If I had to interpret it, I'd say it's simply a random idea or phrase that you jotted down. Without more context, it's difficult to determine what this thought might mean or where it came from.
If you could provide more information about the purpose of the list and the thought, I'd be happy to help further!}

\subsubsection{Example Grasp Stop}
Example of an \eRAG-strict trial in the large environment that was stopped early due to poor grasping (i.e., robot arm was going to collide with object). Task was to ``locate the duck and pick it up."

\texttt{Question: Locate the duck and pick it up.}

\texttt{Thought: I need to find the location of the duck first, then move near it, grab it, and finally release it if necessary. I can use the Locate by ID tool to find the duck's position.}

\texttt{Action: Locate by ID}

\texttt{Action Input: duck$\_$1}

\texttt{Observation:duck:3.40,-0.47, Here is a potentially relevant subset of observed objects so far: ["Floor", "desk$\_$85", "desk$\_$76", "desk$\_$78", "desk$\_$49", "chair$\_$14", "chair$\_$12", "chair$\_$13", "chair$\_$41", "duck$\_$1", "duck$\_$136", "duck$\_$3", "dog$\_$129", "concrete$\_$138", "truss$\_$151", "truss$\_$43", "concrete$\_$153", "shelves$\_$72", "truss$\_$53", "truss$\_$67", "ladder$\_$119", "ladder$\_$127", "ladder$\_$34", "ladder$\_$33"]}

\texttt{I need to move near the duck first, so I can see it clearly and grab it. Then, I'll use the Grab tool to pick it up.}

\texttt{Action: Move Near}

\texttt{Action Input: 3.40,-0.47}

\texttt{Observation:Arrived at '1.95,-0.19,-0.03'. Here is a potentially relevant subset of observed objects so far: ["Floor", "desk$\_$85", "desk$\_$76", "desk$\_$78", "desk$\_$49", "chair$\_$14", "chair$\_$12", "chair$\_$13", "chair$\_$41", "duck$\_$1", "duck$\_$136", "duck$\_$3", "dog$\_$129", "concrete$\_$138", "truss$\_$151", "truss$\_$43", "concrete$\_$153", "shelves$\_$72", "truss$\_$53", "truss$\_$67", "ladder$\_$119", "ladder$\_$127", "ladder$\_$34", "ladder$\_$33"]}

\texttt{I've arrived near the duck, but I still need to grab it using the Grab tool.}

\texttt{Action: Grab}

\texttt{Action Input: The duck}

\texttt{Observation: Grasp service returned: Timed Out. Here is a potentially relevant subset of observed objects so far: ["Floor", "desk$\_$85", "desk$\_$76", "desk$\_$78", "desk$\_$49", "chair$\_$14", "chair$\_$12", "chair$\_$13", "chair$\_$41", "duck$\_$1", "duck$\_$136", "duck$\_$3", "dog$\_$129", "concrete$\_$138", "truss$\_$151", "truss$\_$43", "concrete$\_$163", "shelves$\_$72", "truss$\_$53", "truss$\_$67", "ladder$\_$119", "ladder$\_$127", "ladder$\_$34", "ladder$\_$33"]}

\texttt{It seems like the Grab action timed out, which means the robot was unable to successfully grasp the duck. This might be due to the robot not being able to reach the duck or some other issue preventing it from grabbing it.}

\texttt{Let's try a different approach. Since the robot has arrived near the duck and is unable to grab it, let's use the Ground View and Grab tool instead. This will allow the robot to search for the duck on the ground before trying to grasp it.}

\texttt{Action: Ground View and Grab}

\texttt{Action Input: The duck}

\textbf{Human stop.}

\end{document}